# Nonparametric Bayesian sparse graph linear dynamical systems

**Rahi Kalantari**
Electrical & Computer Engineering
University of Texas at Austin
rkalantari@utexas.edu

**Joydeep Ghosh**
Electrical & Computer Engineering
University of Texas at Austin
jghosh@utexas.edu

**Mingyuan Zhou**
McCombs School of Business
University of Texas at Austin
mzhou@utexas.edu

## Abstract

A nonparametric Bayesian sparse graph linear dynamical system (SGLDS) is proposed to model sequentially observed multivariate data. SGLDS uses the Bernoulli-Poisson link together with a gamma process to generate an infinite dimensional sparse random graph to model state transitions. Depending on the sparsity pattern of the corresponding row and column of the graph affinity matrix, a latent state of SGLDS can be categorized as either a non-dynamic state or a dynamic one. A normal-gamma construction is used to shrink the energy captured by the non-dynamic states, while the dynamic states can be further categorized into live, absorbing, or noise-injection states, which capture different types of dynamical components of the underlying time series. The state-of-the-art performance of SGLDS is demonstrated with experiments on both synthetic and real data.

## 1 INTRODUCTION

Linear dynamical systems (LDSs) are widely used to model multivariate real-valued data that are sequentially observed over time (Ghahramani and Roweis, 1999; Kalman, 1960; Ljung, 1999; West and Harrison, 1997), with highly diverse applications such as financial time series analysis (Carvalho and Lopes, 2007), movement trajectory modeling (Fox et al., 2009), word embedding (Belanger and Kakade, 2015), and sports analytics (Linderman et al., 2017). The LDS assumes that each observation is generated from a multivariate normal distribution whose mean depends on an underlying vector of $K$ latent states. The latent state vectors are represented as continuous random variables related via linear Gaussian dynamics.

Evolving over time under a Markov chain, the latent state vector at time $t$ is produced by adding Gaussian noise to the multiplication of a $K \times K$ state-transition matrix and the latent state vector at time $t-1$. Thus the value of latent state $k$ at time $t$ is a noisy version of the inner product of the $k$th row of the state-transition matrix and the latent state vector at time $t-1$; and the latent state vector at time $t$ is a noisy version of the weighted combination of the columns of the state-transition matrix, with the $k$th column weighted by the value of state $k$ at time $t-1$.

The $K \times K$ state-transition matrix plays an essential role in determining the dynamical behaviors of the LDS. It is often learned from the observed sequences using gradient decent (Ljung, 1999), expectation maximization (EM) (Ghahramani and Hinton, 1996), or subspace identification methods (Van Overschee and De Moor, 2012). The parameters of the LDS state-transition matrix are often regularized during learning (Boots et al., 2008; Liu and Hauskrecht, 2015; Städler et al., 2013). For example, Boots et al. (2008) impose a stability constraint that none of the eigenvalues of the state-transition matrix are greater than one, and Liu and Hauskrecht (2015) impose on the state-transition matrix a low-rank assumption by penalizing the nuclear norm, or a sparse assumption by placing a multivariate Laplacian over the rows.

Despite the tremendous success of the LDS in a wide variety of real applications and the existence of a rich set of algorithms to infer the parameters of the LDS, there is a lack of sound solutions to address the sensitivity of the LDS's performance to the choice of $K$, and even if an appropriate $K$ is identified, the $K \times K$ state-transition matrix could remain difficult to interpret especially if $K$ is not sufficiently small. To address both issues, we propose a nonparametric Bayesian sparse graph LDS (SGLDS) that uses the Bernoulli-Poisson link (Zhou, 2015) to induce sparsity to the state-transition graph affinity matrix, uses the gamma process (Ferguson, 1973) to support $K \to \infty$ in the prior, and uses a normal-gamma construction to shrink the total energy captured by a non-dynamic state, whose corresponding row and column of the state-transition matrix only contain zeros.

The inherent shrinkage mechanism of the gamma process, combined with the Bernoulli-Poisson sparsity-promoting prior on the state-transition matrix, allows the SGLDS to infer in the posterior a parsimonious set of dynamic latent states, each of which has at least one nonzero element in its corresponding row or column of the state-transition matrix. These dynamic latent states are used to capture the underlying dynamical behaviors, where the energy captured by the non-dynamic ones are encouraged to be small by a shrinkage mechanism induced by a normal-gamma construction. Moreover, for latent state $k$ that is active, depending on whether there is at least one nonzero element in both the $k$th row and $k$th column of the state-transition graph affinity matrix, whether there is at least one nonzero element in the $k$th row but no nonzero element in the $k$th column, and whether there is no nonzero element in the $k$th row but at least one nonzero element in the $k$th column, we can further categorize state $k$ as a "live" state, "absorbing" state, and "noise-injection" state, respectively. Decomposing a time series into these latent states, we will show that the live states often capture the trend and seasonal components, while the absorbing and noise-injection states often capture the non-stationary ones.

## 2 NONPARAMETRIC BAYESIAN HIERACHICAL MODEL

The hierarchical model of the LDS is expressed as

$$\begin{aligned} \boldsymbol{y}_t &\sim \mathcal{N}(\mathbf{D}\boldsymbol{x}_t, \boldsymbol{\Phi}^{-1}), \\ \boldsymbol{x}_t &\sim \mathcal{N}(\mathbf{C}\boldsymbol{x}_{t-1}, \boldsymbol{\Lambda}^{-1}), \end{aligned} \quad (1)$$

where $\boldsymbol{y}_t \in \mathbb{R}^P$ and $\boldsymbol{x}_t \in \mathbb{R}^K$ are the observed data vector and latent state vector, respectively, at time $t \in \{1, \ldots, T\}$, $\mathbf{D} = (\boldsymbol{d}_1, \ldots, \boldsymbol{d}_K) \in \mathbb{R}^{P \times K}$ is the observation factor loading matrix, $\mathbf{C} \in \mathbb{R}^{K \times K}$ is the state-transition matrix, and $\boldsymbol{\Phi} \in \mathbb{R}^{P \times P}$ and $\boldsymbol{\Lambda} \in \mathbb{R}^{K \times K}$ are both precision (inverse covariance) matrices.

The learning of the state-transition matrix often needs to be appropriately regularized (Boots et al., 2008; Liu and Hauskrecht, 2015). In this paper, we propose to use a spike-and-slab (Ishwaran and Rao, 2005; Mitchell and Beauchamp, 1988; Zhou et al., 2009) construction for the state-transition matrix, which uses the element-wise product of a real-valued matrix $\mathbf{W} \in \mathbb{R}^{K \times K}$ and a binary matrix $\mathbf{Z} \in \{0,1\}^{K \times K}$ to construct $\mathbf{C}$ as $\mathbf{C} = \mathbf{W} \odot \mathbf{Z}$. We use a conjugate Wishart prior on the precision matrix $\boldsymbol{\Phi}$, let $\boldsymbol{\Lambda} = \text{diag}(\lambda_1, \ldots, \lambda_K)$, and draw $\lambda_k$ from the gamma distribution. We will show how these constructions together with a sparse infinite-dimension $\mathbf{Z}$, generated using the gamma process and the Bernoulli-Poisson link, can be used to infer a parsimonious set of dynamic states to model the dynamics.

### 2.1 Sparse Infinite State-Transition Matrix

To encourage a binary matrix to be sparse and allow $K \to \infty$ in the prior, a common strategy is to use the beta-Bernoulli process or Indian buffet process (Griffiths and Ghahramani, 2005; Thibaux and Jordan, 2007; Zhou et al., 2009). However, here we need to construct a binary matrix $\mathbf{Z}$ that is potentially infinite in both its row and column dimensions, and we hope to express the idea that the probability for $z_{ij} = 1$ is positively associated with both the popularity of state $i$ and that of state $j$. For this reason, we introduce a gamma process $G \sim \Gamma(G_0, 1/c_0)$ defined on the product space $\mathbb{R}^+ \times \Omega$, where $\mathbb{R}^+ = \{x : x > 0\}$, $\Omega$ is a complete separable metric space, $c_0$ is a positive scale parameter, and $G_0$ is a finite and continuous base measure, such that $G(A_j) \sim \text{Gamma}(G_0(A_j), 1/c_0)$ are independent gamma random variables for disjoint partitions $A_j$ of $\Omega$ (Ferguson, 1973). A gamma process draw can be expressed as $G = \sum_{k=1}^{\infty} r_k \delta_{\boldsymbol{d}_k}$, where $\boldsymbol{d}_k \in \mathbb{R}^P$ is an atom, which serves as the $k$th column of $\mathbf{D}$, and $r_k = G(\boldsymbol{d}_k)$ is its weight. We further mark each random point $(r_k, \boldsymbol{d}_k)$ of the gamma process with a gamma random variable as $\lambda_k \sim \text{Gamma}(a, 1/b)$ (Kingman, 1993), which will be shown to help shrink the energy of a non-dynamic state.

Given $\{r_k\}_{1,\infty}$ from the gamma process, we draw the $(i,j)$th element of $\mathbf{Z}$ under the Bernoulli-Poisson link (Zhou, 2015) as

$$z_{ij} = \delta(m_{ij} \geq 1), \ m_{ij} \sim \begin{cases} \text{Pois}(r_i r_j), & \text{if } i \neq j, \\ \text{Pois}(r_0 r_j), & \text{if } i = j, \end{cases} \quad (2)$$

where $m_{ij} \in \mathbb{Z}$, $\mathbb{Z} := \{0, 1 \ldots, \}$, and $\delta(x) = 1$ if the condition $x$ is satisfied and $\delta(x) = 0$ otherwise. With the latent counts $m_{ij}$ marginalized out, we have

$$z_{ij} \sim \begin{cases} \text{Bernoulli}(1 - e^{-r_i r_j}), & \text{if } i \neq j, \\ \text{Bernoulli}(1 - e^{-r_0 r_j}), & \text{if } i = j. \end{cases} \quad (3)$$

The model shown in (2) and (3) is closely related to both the sparse random graph model of Caron and Fox (2015) and the edge partition model of Zhou (2015) in using the Bernoulli-Poisson link to construct a sparse binary matrix with the same numbers of rows and colums. Note we do not impose $z_{ij} = z_{ji}$ for $i \neq j$.

Under the Bernoulli-Poisson link, as discussed in Zhou

(2017), the contribution of $z_{ij}$ to the negative log-likelihood of the model can be expressed as

$$-z_{ij} \ln[1 - \exp(-2r_i r_j)] + 2(1 - z_{ij})r_i r_j,$$

which quickly explodes towards $\infty$ as the product $r_i r_j$ approaches zero when $z_{ij} = 1$. Thus the choice of this link function implies a strong penalty for $z_{ij} = 1$ when its corresponding product $r_i r_j$ is small, while not strongly penalizing $z_{ij} = 0$ when $r_i r_j$ is large. Denoting $\gamma_0 = G_0(\Omega)$ as the mass parameter of the gamma process, below we further show that the total number of nonzero elements in $\mathbf{Z}$ is finite in expectation.

**Lemma 1.** *The expectation of the total number of nonzero elements in $\mathbf{Z}$, denoted as $\|\mathbf{Z}\|_0 = \sum_{i=1}^{\infty} \sum_{j=1}^{\infty} z_{ij}$, is bounded as follows*

$$\mathbb{E}[\|\mathbf{Z}\|_0] \leq r_0 \frac{\gamma_0}{c_0} + \frac{\gamma_0^2}{c_0^2}. \quad (4)$$

### 2.2 Sparse Graph Linear Dynamical System

The gamma process $G \sim \text{Gamma}(G_0, 1/c_0)$ has an inherent shrinkage mechanism, as the number of atoms with weights greater than $\epsilon > 0$ follows $\text{Pois}(\gamma_0 \int_\epsilon^\infty r^{-1} e^{-c_0 r} dr)$. For the convenience of implementation, we truncate the total number of atoms at $K$ by choosing a finite and discrete base measure as $G_0 = \sum_{k=1}^{K} \frac{\gamma_0}{K} \delta_{(\lambda_k, \boldsymbol{d}_k)}$. The hierarchical model of the (truncated) sparse graph LDS can be expressed as

$$\boldsymbol{y}_t \sim \mathcal{N}(\mathbf{D}\boldsymbol{x}_t, \boldsymbol{\Phi}^{-1}), \quad \boldsymbol{\Phi} \sim \text{Wishart}(\mathbf{V}, P+2),$$
$$\boldsymbol{x}_t \sim \mathcal{N}\left[(\mathbf{W} \odot \mathbf{Z})\boldsymbol{x}_{t-1}, \text{diag}(\lambda_1, \ldots, \lambda_K)^{-1}\right],$$
$$\lambda_k \sim \text{Gamma}(a, 1/b), \quad \boldsymbol{d}_k \sim \mathcal{N}\left(\mathbf{0}, \mathbf{I}_P/\sqrt{P}\right),$$
$$w_{ij} \sim \mathcal{N}(0, \varphi_{ij}^{-1}), \quad \varphi_{ij} \sim \text{Gamma}(\alpha_0, 1/\beta_0),$$
$$z_{ij} = \delta(m_{ij} \geq 1), \quad m_{ij} \sim \begin{cases} \text{Pois}(r_i r_j), & \text{if } i \neq j, \\ \text{Pois}(r_0 r_j), & \text{if } i = j, \end{cases}$$
$$r_k \sim \text{Gamma}(\gamma_0/K, 1/c_0), \quad \gamma_0 \sim \text{Gamma}(a_0, 1/b_0),$$
$$c_0 \sim \text{Gamma}(a_0, 1/b_0), \quad \boldsymbol{x}_0 \sim \mathcal{N}(\boldsymbol{m}_0, \mathbf{H}_0). \quad (5)$$

Where $\boldsymbol{d}_k$ is $k$th column of matrix $\mathbf{D}$, $z_{ij}$ is the $(i,j)$th element of matrix $\mathbf{Z}$ and $\boldsymbol{\Lambda} = \text{diag}(\lambda_1, \ldots, \lambda_K)$. A graphical representation of the generative model is shown in the Appendix.

As in Lemma 1, the total number of nonzero elements in $\mathbf{Z}$ has a finite expectation. Thus if the gamma process truncation level $K$ is set large enough, it is expected for some state $i$ that $\sum_k z_{ik} = 0$, which means its corresponding row in $\mathbf{Z}$ has no nonzero elements, and/or $\sum_k z_{ki} = 0$, which means its corresponding column in $\mathbf{Z}$ has no nonzero elements. Note that

$$x_{ti} \sim \mathcal{N}\left(\sum_k w_{ik} z_{ik} x_{(t-1)k}, \lambda_i^{-1}\right),$$

where $x_{ti}$ is the $i$th element of $\boldsymbol{x}_t = (x_{t1}, \ldots, x_{tK})'$, and

$$\boldsymbol{x}_{t+1} \sim \mathcal{N}\left[\sum_i (\boldsymbol{w}_i \odot \boldsymbol{z}_i) x_{ti}, \text{diag}(\lambda_1, \ldots, \lambda_K)^{-1}\right],$$

where $\boldsymbol{w}_i$ and $\boldsymbol{z}_i$ are the $i$th columns of $\mathbf{W}$ and $\mathbf{Z}$, respectively. Thus depending on whether $\sum_k z_{ik} = 0$ and/or $\sum_k z_{ki} = 0$, we can categorize state $k$ as one of the four different types of states:

- **Live state**: if both $\sum_k z_{ik} > 0$ and $\sum_k z_{ki} > 0$, which means both a nonzero row and a nonzero column for state $i$, then $x_{ti}$, the value of state $i$ at time $t$, not only is influenced by $\sum_k z_{ik}$ latent states of the previous time, but also influences $\sum_k z_{ki}$ states of the next time; we find a live state often captures trend and seasonal components.

- **Absorbing state**: if $\sum_k z_{ik} > 0$ but $\sum_k z_{ki} = 0$, which means a nonzero row but a zero column for state $i$, then $x_{ti}$ is influenced by $\sum_k z_{ik}$ latent states of the previous time, but since

$$\boldsymbol{x}_{t+1} \sim \mathcal{N}\left[\sum_{i' \neq i}(\boldsymbol{w}_{i'} \odot \boldsymbol{z}_{i'}) x_{ti'}, \text{diag}(\lambda_1, \ldots, \lambda_K)^{-1}\right],$$

$x_{ti}$ does not influence the values of the latent states of the next time. We show an absorbing state often catches some dynamical components not captured by the live states, as well as some static components of the time series.

- **Noise-injection state**: if $\sum_k z_{ik} = 0$ but $\sum_k z_{ki} > 0$, which means a zero row but a nonzero column for state $i$, then we have

$$x_{ti} \sim \mathcal{N}\left(0, \lambda_i^{-1}\right),$$

which means that $x_{ti}$ does not dependent on the values of the latent states at the previous time, but influences $\sum_k z_{ki}$ states of the next time.

- **Non-dynamic state**: if both $\sum_k z_{ik} = 0$ and $\sum_k z_{ki} = 0$, then $x_{ti}$ captures only the static noise component of the data. Moreover, due to the normal-gamma construction, we have

$$P(x_{1i}, \ldots, x_{Ti} \,|\, a, b)$$
$$= \int_0^\infty \left[\prod_{t=1}^T \mathcal{N}(x_{ti}; 0, \lambda_i^{-1})\right] \text{Gamma}(\lambda_i; a, 1/b) \, d\lambda_i$$
$$= \frac{b^a \Gamma(a + \frac{T}{2})}{(2\pi)^{\frac{T}{2}} \Gamma(a)} \left(b + \frac{1}{2} \sum_{t=1}^T x_{ti}^2\right)^{-(a + \frac{T}{2})}, \quad (6)$$

which provides a strong shrinkage for $\sum_{t=1}^T x_{ti}^2$ that represents the total energy captured by state $i$. Note that when $T = 1$, this shrinkage mechanism becomes the same as the one used in Tipping (2001) for obtaining sparse solutions to regression and classification tasks. Since $T > 1$ due to the nature of time series data, the normal-gamma construction used here provides an even stronger shrinkage than the one used in Tipping (2001).

Note without loss of generality, with $z_i := \sum_k (z_{ik} + z_{ki})$, we can rewrite the model by marginalizing out all non-dynamic states as

$$\boldsymbol{y}_t \sim \mathcal{N}\left(\tilde{\mathbf{D}}\tilde{\boldsymbol{x}}_t, \boldsymbol{\Phi}^{-1} + \sum_{i:\, z_i=0} \lambda_i \boldsymbol{d}_i \boldsymbol{d}_i^T\right)$$
$$\tilde{\boldsymbol{x}}_t \sim \mathcal{N}\left[(\tilde{\mathbf{W}} \odot \tilde{\mathbf{Z}})\tilde{\boldsymbol{x}}_{t-1},\ \mathrm{diag}(\tilde{\boldsymbol{\lambda}})^{-1}\right],$$

where $\tilde{\mathbf{D}} = \{\boldsymbol{d}_i\}_{i:z_i>0}$, $\tilde{\boldsymbol{x}}_t = \{\boldsymbol{x}_{ti}\}_{i:z_i>0}$, $\tilde{\mathbf{W}} = \{w_{ij}\}_{ij:z_i>0,z_j>0}$, $\tilde{\mathbf{Z}} = \{z_{ij}\}_{ij:z_i>0,z_j>0}$, and $\tilde{\boldsymbol{\lambda}} = \{\lambda_i\}_{i:z_i>0}$. In addition, since

$$(\lambda_i \,|\, z_i = 0, -) \sim \mathrm{Gamma}\left(a + \frac{T}{2}, \frac{1}{b + \frac{1}{2}\sum_{t=1}^T x_{ti}^2}\right)$$

and the normal-gamma construction encourages the total energy $\sum_{t=1}^T x_{ti}^2$ of a non-dynamic state to be small, as shown in (6), we expect $\lambda_i$ for $i \in \{i : z_i = 0\}$ to be shrunk in their posterior and hence the observation noise covariance to be mainly captured by $\boldsymbol{\Phi}^{-1}$, which has an inverse-Wishart prior as in (5).

### 2.3 Bayesian Inference via Gibbs sampling

1. We sample $\boldsymbol{x}_t$ as

$$(\boldsymbol{x}_t \,|\, -) \sim \mathcal{N}(\boldsymbol{\theta}_t, \boldsymbol{\Sigma_t}),$$

where $\boldsymbol{\Sigma}_T = (\mathbf{D}'\boldsymbol{\Phi}\mathbf{D} + \boldsymbol{\Lambda})^{-1}$, $\boldsymbol{\theta}_T = \boldsymbol{\Sigma}_T(\mathbf{D^T}\boldsymbol{\Phi}\mathbf{y}_T + \boldsymbol{\Lambda}(\mathbf{W}\odot\mathbf{Z})\boldsymbol{x}_{T-1})$, $\boldsymbol{\Sigma_0} = (\mathbf{H_0} + (\mathbf{W}\odot\mathbf{Z})^T\boldsymbol{\Lambda}(\mathbf{W}\odot\mathbf{Z}))^{-1}$, $\boldsymbol{\theta}_0 = \boldsymbol{\Sigma}_0(\mathbf{H_0}\boldsymbol{m}_0 + (\mathbf{W}\odot\mathbf{Z})^T\boldsymbol{\Lambda}\boldsymbol{x}_1)$, and if $1 \leq t \leq T-1$, $\boldsymbol{\Sigma}_t = (\mathbf{D^T}\boldsymbol{\Phi}\mathbf{D} + \boldsymbol{\Lambda} + (\mathbf{W}\odot\mathbf{Z})^T\boldsymbol{\Lambda}(\mathbf{W}\odot\mathbf{Z}))^{-1}$ and $\boldsymbol{\theta}_t = \boldsymbol{\Sigma}_t(\mathbf{D^T}\boldsymbol{\Phi}\mathbf{y}_t + \boldsymbol{\Lambda}(\mathbf{W}\odot\mathbf{Z})\boldsymbol{x}_{t-1} + (\mathbf{W}\odot\mathbf{Z})^T\boldsymbol{\Lambda}\boldsymbol{x}_{t+1})$.

2. We sample $w_{ij}$ as

$$(w_{ij} \,|\, -) \sim \mathcal{N}(\mu_{ij}, \tau_{ij}),$$

where $\tau_{ij} = (z_{ij}\lambda_i T_j + \sigma_{ij})^{-1}$, $\mu_{ij} = \tau_{ij}(z_{ij}\lambda_i Q_{ij})$, $T_j = \sum_{t=1}^T x_{j(t-1)}^2$, $Q_{ij} = \sum_{t=1}^T x_{it}^{-j} x_{j(t-1)}$, and $x_{it}^{-j} = x_{it} - \sum_{k'=1, k'\neq j}^K (w_{ik'} z_{ik'} x_{k'(t-1)})$.

3. We sample $z_{ij}$ as

$$(z_{ij} \,|\, -) \sim \mathrm{Bernoulli}[p_{ij1}/(p_{ij1} + p_{ij0})],$$

where $p_{ij1} = e^{-\frac{1}{2}(w_{ij}^2 T_j \lambda_i - 2w_{ij}\lambda_i Q_{ij})}(1 - e^{-r_i r_j})$ and $p_{ij0} = e^{-r_i r_j}$, where $r_i r_j$ are replaced with $r_0 r_i$ if $i = j$.

4. We sample $m_{ij}$ as

$$(m_{ij} \,|\, -) \sim \begin{cases} z_{ij}\mathrm{Pois}_+(r_i r_j), & \text{if } i \neq j \\ z_{ij}\mathrm{Pois}_+(r_0 r_i), & \text{if } i = j \end{cases}$$

where $x \sim \mathrm{Pois}_+(\lambda)$ is a truncated Poisson distribution with $P(x = k) = (1 - e^{-\lambda})^{-1}\lambda^k e^{-\lambda}/k!$ for $k \in \{1, 2, 3, ...\}$.

5. We sample $r_i$ as

$$(r_i \,|\, -) \sim \mathrm{Gamma}\left(\frac{\gamma_0}{K} + \sum_{j=1}^K m_{ij}, \frac{1}{c_0 + \sum_{j=0, j\neq i}^K r_j}\right).$$

6. Sample $\gamma_0$: using a data augmentation technique for the negative binomial distribution (Zhou and Carin, 2015), we first sample a Chinese restaurant table (CRT) random variable and then sample $\gamma_0$ as

$$(l_i \,|\, -) \sim \mathrm{CRT}\left(\sum_{j=1}^K m_{ij}, \frac{\gamma_0}{K}\right),$$
$$(\gamma_0 \,|\, -) \sim \mathrm{Gamma}\left(a_0 + \sum_{i=1}^K l_i, \frac{1}{b_0 + \sum_{i=1}^K p_i'/K}\right),$$

where $p_i' = -\ln(1 - \frac{s_i}{s_i + c_0})$ and $s_i = \sum_{j=0, j\neq i}^K r_j$.

7. We sample $c_0$ as

$$(c_0 | -) \sim \mathrm{Gamma}\left(\gamma_0 + a_0, \frac{1}{b_0 + \sum_{j=0}^K r_j}\right)$$

8. Sample $\boldsymbol{d}_k$: we sample $\boldsymbol{d}_k$, the $k$th column of $\mathbf{D}$, as

$$(\boldsymbol{d}_k \,|\, -) \sim \mathcal{N}(\boldsymbol{m}_k, \mathbf{E}_k),$$

where $\mathbf{E}_k = (F_k^1 \boldsymbol{\Phi} + \frac{\mathbf{I}_P}{\sqrt{P}})^{-1}$, $\boldsymbol{m}_k = \mathbf{E}_k\boldsymbol{\Phi}(F_k^2 - F_k^3)$, $F_k^1 = \sum_{t=1}^T x_{kt}^2$, $F_k^2 = \sum_{t=1}^T x_{kt}\boldsymbol{y}_t$, and $F_k^3 = \sum_{t=1}^T x_{kt}\mathbf{D}^{-\mathbf{k}}\boldsymbol{x}_t^{-k}$. Here $\mathbf{D}^{-\mathbf{k}}$ refers to matrix $\mathbf{D}$ without the $k$th column, and $\boldsymbol{x}_t^{-k}$ refers to $\boldsymbol{x}_t$ without the $k$th element.

9. We sample $\boldsymbol{\Phi}^{-1}$ as

$$(\boldsymbol{\Phi}^{-1} \,|\, -) \sim \mathrm{InverseWishart}(G + \mathbf{V}, P + 2 + T),$$

where $G = G_1 - G_2 + G_3 - G_4$, $G_1 = \sum_{t=1}^T \boldsymbol{y}_t\boldsymbol{y}_t^T$, $G_2 = \sum_{t=1}^T \boldsymbol{y}_t\boldsymbol{x}_t^T\mathbf{D}^T$, $G_3 = \sum_{t=1}^T \mathbf{D}\mathbf{x_t}\boldsymbol{x}_t^T\mathbf{D}^T$, and $G_4 = \sum_{t=1}^T \mathbf{D}\boldsymbol{x}_t\mathbf{y_t^T}$.

10. We sample $\lambda_i$ as

$$(\lambda_i \,|\, -) \sim \mathrm{Gamma}\Big(a + T/2,$$
$$\frac{1}{b + \sum_{t=1}^T (x_{it} - (W \odot Z)_i \boldsymbol{x}_{t-1})^2/2}\Big),$$

where $(W \odot Z)_i$ is the $i$th row of $\mathbf{W} \odot \mathbf{Z}$.

11. We sample $\varphi_{ij}$ as

$$(\varphi_{ij} \,|\, -) = \mathrm{Gamma}\big(\alpha_0 + 1/2, 1/(w_{ij}^2/2 + \beta_0)\big).$$

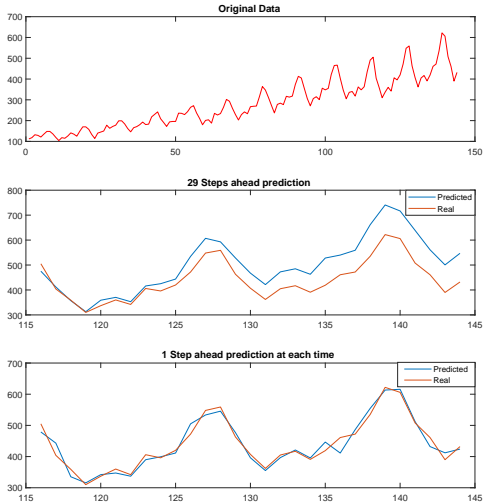

Figure 1: Monthly Airline Passenger Numbers 1949-1960. Top: the original time series. Middle: the predicted time series of the next 29 months given the first 115 months. Bottom: the one-month-ahead prediction given the global parameters learned on the first 115 months and the updated latent state vector of the previous month.

### 2.4 Illustration and Model Interpretation

A time series can often be decomposed into trend, seasonal, auto-regressive, and noise components (West and Harrison, 1997). As a simple illustration, we first apply our sparse graph linear dynamical system (SGLDS) to a one-dimensional dataset for Monthly Airline Passenger Numbers 1949-1960 (Box and Jenkins, 1976). As in Figure 1, the times series exhibits a clear trend and a multiplicative seasonal pattern. We train SGLDS on the first 115 months to obtain the global parameters $\mathbf{D}$, $\mathbf{W}$, $\mathbf{Z}$, $\mathbf{\Phi}$, and $\boldsymbol{\lambda}$, and predict the next 29 months given these global parameters. We first consider a 29 step-ahead-prediction, in which we propagate $\boldsymbol{x}_{115}$ into the future to predict $\boldsymbol{x}_t$ and $\boldsymbol{y}_t$ for $t = 116, \ldots, 144$ given the global parameters, where $\boldsymbol{x}_t$ will not be updated. We then consider one-step-ahead predictions, in which we update $\boldsymbol{x}_t$ given $\boldsymbol{y}_t$ and the global parameters and use the updated $\boldsymbol{x}_t$ to predict $\boldsymbol{x}_{t+1}$ and $\boldsymbol{y}_{t+1}$. As shown in Figure 1, SGLDS well captures the trend, seasonality, and multiplicative nature of the time series.

We apply SGLDS to the Beijing meteorological data of Liang et al. (2015), which consists of the hourly reports of six measurements, including dew point, temperature, pressure (hPa), cumulated wind speed (m/s), cumulated hours of snow, and cumulated hours of rain, in Beijing for the time period from January 1st, 2010 to December 31st, 2014. We use the weekly average of each measurement to train the models. We set the truncation level to be $K = 40$ and display the inferred $40 \times 40$ sparse state-transition matrix $\mathbf{Z}$ in the left part of Figure 2, using which we further draw the corresponding state-transition graph in the right part of Figure 2. It is clear from Figure 2 that while we set the truncation level at $K = 40$, SGLDS infers 24 dynamic states, each of which has at least one nonzero element in the corresponding row and/or column of the inferred state-transition matrix $\mathbf{Z}$.

Motivated by the analysis in Section 2.2, we decompose the reconstruction $\hat{\boldsymbol{y}}_t$ into the superposition of four different times series as

$$\begin{cases} \hat{\boldsymbol{y}}_t^{live} &= \sum_{i:\sum_k z_{ik}>0, \sum_k z_{ki}>0} \boldsymbol{d}_i x_{ti}, \\ \hat{\boldsymbol{y}}_t^{absorbing} &= \sum_{i:\sum_k z_{ik}>0, \sum_k z_{ki}=0} \boldsymbol{d}_i x_{ti}, \\ \hat{\boldsymbol{y}}_t^{noise-injection} &= \sum_{i:\sum_k z_{ik}=0, \sum_k z_{ki}>0} \boldsymbol{d}_i x_{ti}, \\ \hat{\boldsymbol{y}}_t^{non-dynamic} &= \sum_{i:\sum_k z_{ik}=0, \sum_k z_{ki}=0} \boldsymbol{d}_i x_{ti}, \end{cases}$$

which are displayed in rows one to four of Figure 3, respectively. Each column of Figure 3 corresponds one of the six dimensions of the original time series. The original observations and the reconstructed components are shown in red and blue, respectively. It is clear that the live states together capture most of the dynamic components, seasonal ones in particular, but with some artifacts in reconstructing the original time series. For example, the measurements in dimension 1 and 2 tend to be slightly underestimated, specifically in dimension 2 the level of underestimation is smaller than dimension 1, and in dimensions 5 and 6 the live states reconstruct the correct period but wrong phase. Despite these artifacts, it is interesting to notice how the absorbing states help compensate these artifacts in dimensions 1, 2, 4, and 5. It is worth mentioning that in the last three dimensions, for regularly spaced spikes, the model well captures their temporal locations but not amplitudes, whereas for spikes appearing in random temporal locations, the model captures neither their locations nor amplitudes. For these data irregularities (sudden spikes with random amplitudes and/or locations), we don't expect our model to capture them. In fact, we would worry about overfitting if a model does capture these irregular temporal behaviors. In addition, as it was expected, the noise-injection and non-dynamic states play little role in reconstructing the observations at time $t$. However, we find that the noise-injection states do play a clear role in influencing observations at time $t' > t$ with the noise they injected at time $t$, which may trigger not only short-term non-stationary sequences, but also long-term seasonable components (not shown here due to space constraint).

To further illustrate how the live states capture seasonal components, we mark three representative loops in Figure 2 with different colors, and for each loop, we plot in Figure C.3 of the Appendix the data reconstructed using only the states belonging to that loop.

Nonparametric Bayesian sparse graph linear dynamical systems

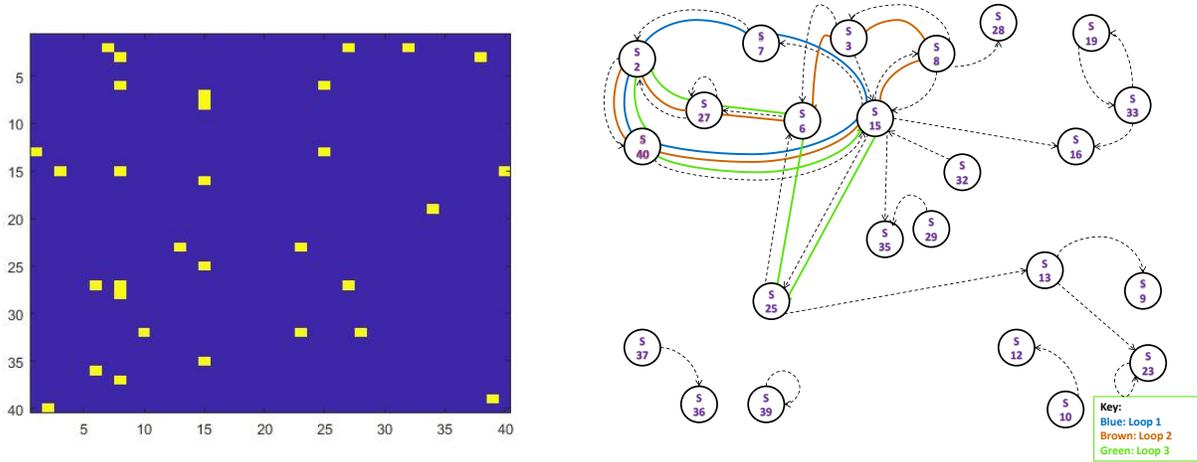

Figure 2: Left: The inferred sparse state-transition graph affinity matrix of SGLDS on the Beijing meteorological data $\mathbf{Z}$, where the truncation level is set as $K = 40$. Right: The corresponding state transition graph.

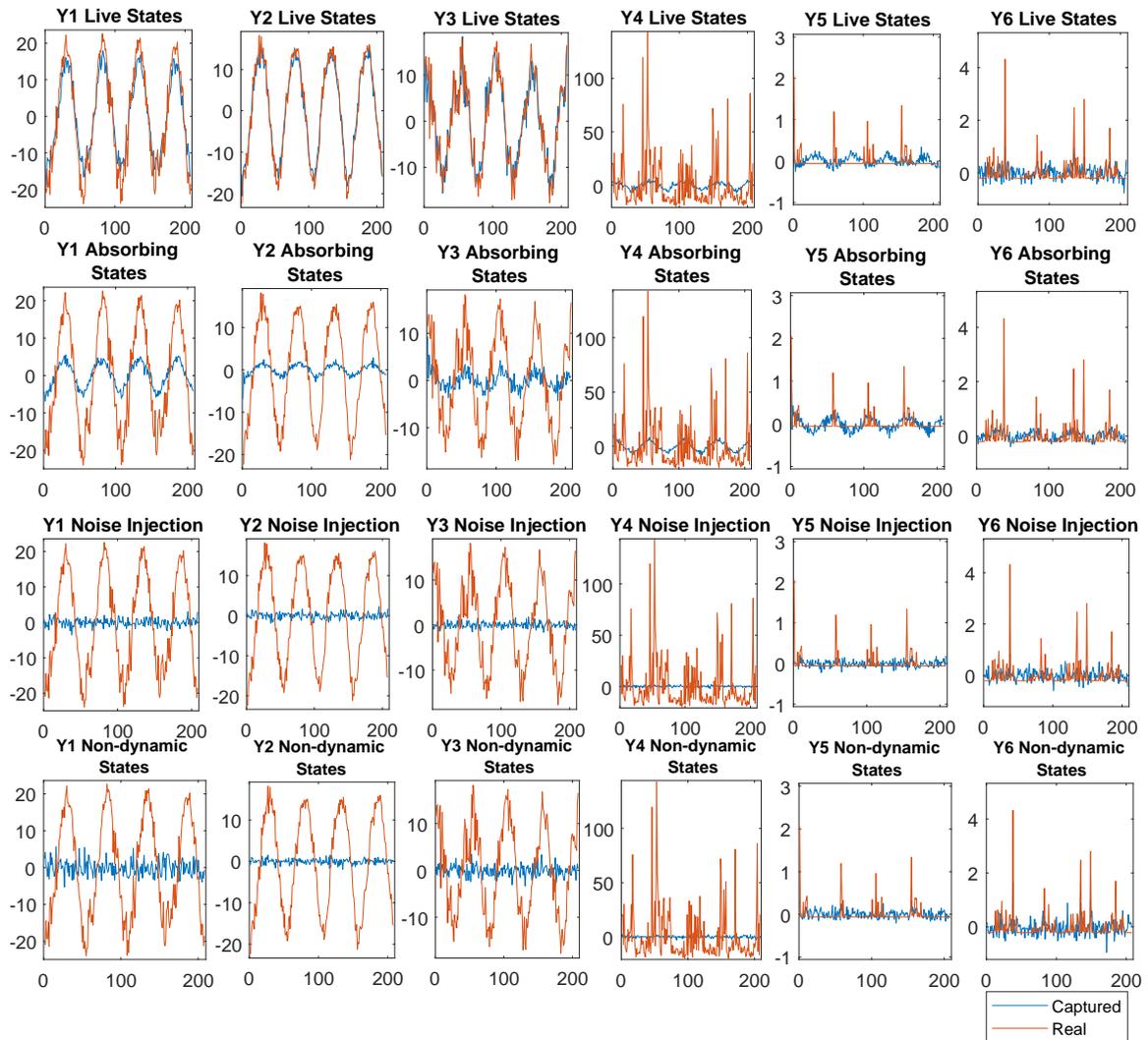

Figure 3: The six-dimensional time series of the Beijing meteorological data and the reconstructed components using the live, absorbing, noise-injection, or non-dynamic states.

Note that the states of different loops could overlap, and hence the reconstructed loop-dependent times series could share dynamical components. As clearly shown in Figure C.3, these three different (overlapping) loops capture somewhat different seasonal components. For example, the Green loop captures the correct period and phase of the time series in Dimension 3, whereas loops 1 and 2 capture the correct period but shifted phase.

## 3 EXPERIMENTAL RESULTS

We use both synthetic and real datasets to further verify the properties and performance of the proposed nonparametric Bayesian SGLDS. Note that as our nonparametric Bayesian algorithm is able to infer the number of dynamic states from the data and shrink the energies captured by non-dynamic states, the performance of SGLDS is not sensitive to the setting of the truncation level $K$ as long as $K$ is sufficiently large. In addition, our Gibbs sampling based inference is not sensitive to initialization. In fact, we achieve the state-of-the-art performance, as reported below, by simply initializing the parameters of SGLDS at random. Moreover, due to the generative nature of the hierarchical Bayesian model, SGLDS can easily handle missing data and capture the underlying dynamics given only a short sequence of observations.

To measure the predictive performance of a dynamical system, we consider both the squared error as

$$\text{SE} = \sqrt{\sum_{t=1}^{T}\sum_{p=1}^{P}(y_{td}-\hat{y}_{td})^2},$$

which is the Frobenius norm of the difference between the predictions and observations, and the average mean absolute percentage error (AMAPE) used in Liu and Hauskrecht (2015), which is defined as

$$\text{AMAPE} = \frac{1}{TP}\sum_{t=1}^{T}\sum_{p=1}^{P}\left|1-\left|\frac{\hat{y}_{dt}}{y_{dt}}\right|\right|.$$

AMAPE measures the proportion of deviation relative to the true values.

We consider three different baselines for comparison, including 1) LDS inferred with EM Ghahramani and Hinton (1996), 2) rLDS$_G$: a regularized LDS of Liu and Hauskrecht (2015) that uses the $L_1$ penalty to remove the unnecessary rows of the state-transition matrix, and 3) rLDS$_R$: a regularized LDS of Liu and Hauskrecht (2015) that penalizes the nuclear norm of the state-transition matrix to achieve a low rank solution.

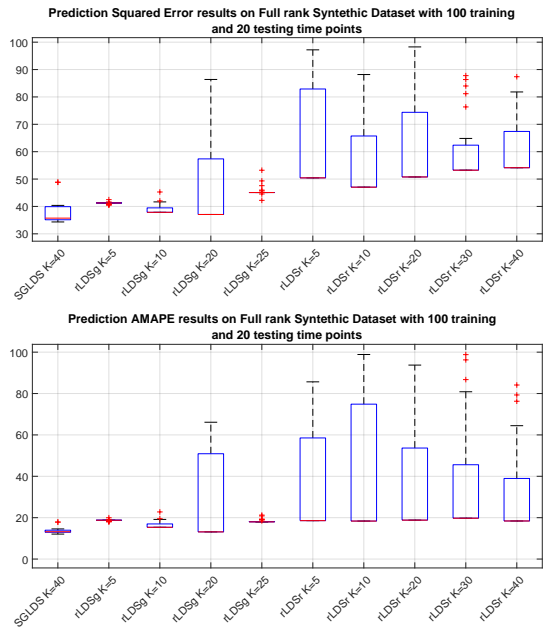

Figure 4: Comparison of the prediction performance of different algorithms on a 12 dimensional synthetic data generated with a random non-sparse state-transition matrix.

### 3.1 Performance Evaluation on Datasets

First, a synthetic time series of dimension $P = 12$ and length $T = 120$ is generated from a LDS with a $10 \times 10$ state-transition matrix $\mathbf{C}$, whose independent, and identically distributed ($i.i.d.$) elements are drawn from $\mathcal{N}(0.1, 0.2)$, and a random observation matrix $\mathbf{D}$, whose $i.i.d.$ columns are drawn from $\mathcal{N}(\mathbf{0}, \mathbf{I}_P)$. The other parameters of the LDS are specified as $\mathbf{\Phi} = 0.1\mathbf{I}_P$ and $\mathbf{\Lambda} = 0.5\mathbf{I}_{10}$, the state vector is initialized with $\boldsymbol{x}_0 \sim \mathcal{N}(\mathbf{1}, \mathbf{I}_{10})$. The models are trained using the first 100 observations and the last 20 ones are used to measure the prediction performance. We consider one-step-ahead predictions, during which the inferred global model parameters, including the observation, state-transition, and covariance matrices, will be fixed; to make prediction for $\boldsymbol{y}_{t+1}$ for $t \geq 100$, we first update the local parameter $\boldsymbol{x}_t$ given $\boldsymbol{y}_t$ and the global parameters and then use the updated $\boldsymbol{x}_t$ to predict $\boldsymbol{x}_{t+1}$ and then $\boldsymbol{y}_{t+1}$. The average sparsity levels and ranks of the transition matrices inferred under rLDS$_G$, and rLDS$_R$, and SGLDS will be reported. We consider 25 random trials for each algorithm and show the prediction performance in Figure 4. Note SGLDS is a (truncated) Bayesian nonparametric model, for which we set the truncation level as large as $K = 40$, which is large enough to accommodate all dynamic states for the times series considered in the paper. For the SGLDS, the average percentage of zero elements of the truncated $40 \times 40$ state-transition-matrix over different runs is as high as 98.2%, and consequently, the average rank is as low as 9.3. By contrast, rLDS$_r$ has an average rank of 30.7 when $K = 40$ and rLDS$_G$ has average rank of 9.4 when $K = 20$, while having large performance variations both across different random trials and with different $K$, as shown in Figure 4.

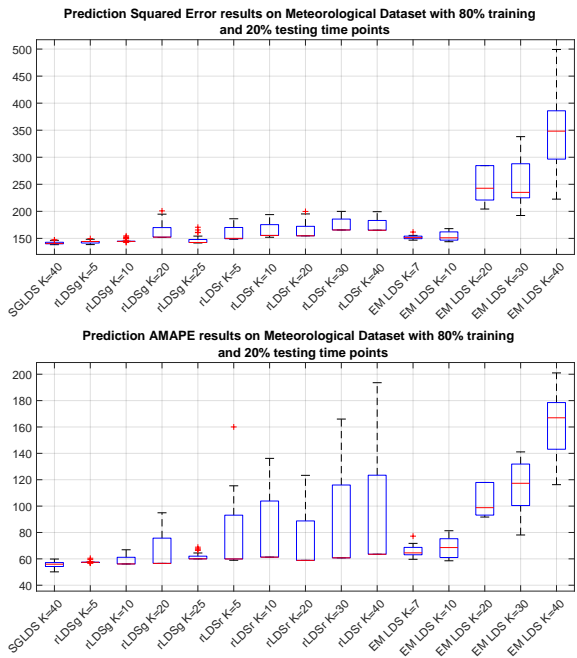

Figure 5: Analogous plots to these in Figure 4 for the Beijing meteorological data.

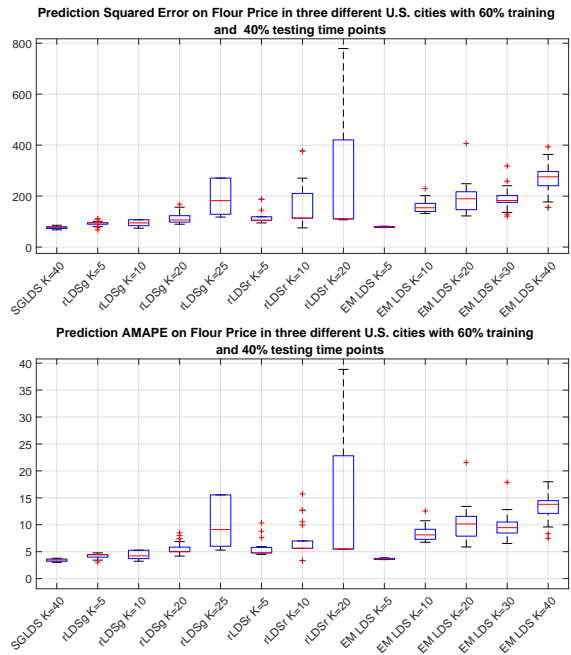

Figure 6: Analogous plots to these in Figure 4 for the flour price data.

Second, we consider the Beijing meteorological data, a six dimensional data used for illustration in Section 2.4, to train and test the models. We use the first 80% of the data for training and the remaining 20% for testing. To use AMAPE as a meaningful metric for this dataset, we use the first 4 dimensions of the time series to compute AMAPE, as the majority of the observations in dimensions 5 and 6 are zeros. Given a fixed $K$, the parameters of the rLDS$_G$ and rLDS$_R$ models are tuned by cross validation. As shown in Figure 5, the proposed SGLDS not only outperforms the other models, but also has very stable results across different runs. For the inferred truncated $40 \times 40$ state-transition matrix, the average proportion of zeros is 84% and average rank is 25.2 over different runs. By contrast, the average rank of rLDS$_G$ is 14.8 and that of rLDS$_R$ is 18.5 when $K = 20$, with which both rLDS$_G$ and rLDS$_R$ have their best performance. It is clear that the EM algorithm overfits the data as $K$ becomes large, resulting in poor prediction performance.

Another real dataset considered in the paper is the monthly flour price data in three different cities of the United States from August 1972 to November 1980 (Reinsel, 2003). We use the first 60% of the observations for training and the remaining 40% for testing. Given a fixed $K$, the other parameters of the rLDS$_G$ and rLDS$_R$ are again optimized by cross validation. Not surprisingly, as shown in Figure 6, SGLDS outperforms all the other models. For the inferred truncated $40 \times 40$ state-transition matrix, the average sparsity level is about 97.1% and average rank is 11.1 over different runs. The average rank of rLDS$_R$ is 17.8 when $K = 20$ and average rank of rLDS$_G$ is 13.5 when $K = 25$. Note if imposing a stronger regularization to obtain a lower-rank solution, the performance of rLDS often degrades significantly. Note the results of rLDS$_R$ with $K = 30$ and $K = 40$ are not shown due to their poor performance.

## 4 CONCLUSIONS

We integrate the gamma process, Bernoulli-Poisson link, and normal-gamma construction into a single generative model, leading to a novel nonparametric Bayesian sparse graph linear dynamical system (SGLDS). The proposed model sidesteps the need to tune the size of the latent state-transition matrix, and infers a finite number of dynamic states that can be further categorized into live, absorbing, and noise-injection states, which capture different types of dynamical behaviors of the data. The spares random graph, whose affinity matrix is the inferred sparse state-transition matrix, leads to a clear interpretation of the inferred model parameters. Experiments on both synthetic and real data show that SGLDS not only achieves state-of-the-art performance, but also infers clearly interpretable latent structures.

# References

D. Belanger and S. Kakade. A linear dynamical system model for text. In *International Conference on Machine Learning*, pages 833–842, 2015.

B. Boots, G. J. Gordon, and S. M. Siddiqi. A constraint generation approach to learning stable linear dynamical systems. In *NIPS*, pages 1329–1336, 2008.

G. E. Box and G. M. Jenkins. Time series analysis. forecasting and control. In *Holden-Day Series in Time Series Analysis, Revised ed., San Francisco: Holden-Day, 1976*, 1976.

F. Caron and E. B. Fox. Sparse graphs using exchangeable random measures. *arXiv:1401.1137v3*, 2015.

C. M. Carvalho and H. F. Lopes. Simulation-based sequential analysis of Markov switching stochastic volatility models. *Computational Statistics & Data Analysis*, 51(9):4526–4542, 2007.

T. S. Ferguson. A Bayesian analysis of some nonparametric problems. *Ann. Statist.*, 1(2):209–230, 1973.

E. Fox, E. B. Sudderth, M. I. Jordan, and A. S. Willsky. Nonparametric bayesian learning of switching linear dynamical systems. In *NIPS*, pages 457–464. 2009.

Z. Ghahramani and G. E. Hinton. Parameter estimation for linear dynamical systems. Technical report, 1996.

Z. Ghahramani and S. T. Roweis. Learning nonlinear dynamical systems using an EM algorithm. In *NIPS*, pages 431–437, 1999.

T. L. Griffiths and Z. Ghahramani. Infinite latent feature models and the Indian buffet process. In *NIPS*, 2005.

H. Ishwaran and J. S. Rao. Spike and slab variable selection: frequentist and Bayesian strategies. *Annals of statistics*, pages 730–773, 2005.

R. E. Kalman. A new approach to linear filtering and prediction problems. *Transactions of the ASME–Journal of Basic Engineering*, 82(Series D):35–45, 1960.

J. F. C. Kingman. *Poisson Processes*. Oxford University Press, 1993.

X. Liang, T. Zou, B. Guo, S. Li, H. Zhang, S. Zhang, H. Huang, and S. X. Chen. Assessing beijing's pm2.5 pollution: severity, weather impact, apec and winter heating. In *Proc. R. Soc. A*, volume 471, page 20150257. The Royal Society, 2015.

S. Linderman, M. Johnson, A. Miller, R. Adams, D. Blei, and L. Paninski. Bayesian learning and inference in recurrent switching linear dynamical systems. In *AISTATS*, volume 54, pages 914–922, Fort Lauderdale, FL, USA, 20–22 Apr 2017.

Z. Liu and M. Hauskrecht. A regularized linear dynamical system framework for multivariate time series analysis. In *AAAI*, pages 1798–1804, 2015.

L. Ljung. *System Identification: Theory for the User, 2nd edition*. Prentice Hall, 1999.

T. J. Mitchell and J. J. Beauchamp. Bayesian variable selection in linear regression. *Journal of the American Statistical Association*, 83(404):1023–1032, 1988.

G. C. Reinsel. *Elements of multivariate time series analysis*. Springer Science & Business Media, 2003.

N. Städler, S. Mukherjee, et al. Penalized estimation in high-dimensional hidden markov models with state-specific graphical models. *The Annals of Applied Statistics*, 7(4):2157–2179, 2013.

R. Thibaux and M. I. Jordan. Hierarchical beta processes and the Indian buffet process. In *AISTATS*, 2007.

M. Tipping. Sparse Bayesian learning and the relevance vector machine. *J. Mach. Learn. Res.*, 1: 211–244, June 2001.

P. Van Overschee and B. De Moor. *Subspace identification for linear systems: Theory – Implementation – Applications*. Springer Science & Business Media, 2012.

M. West and J. Harrison. *Bayesian Forecasting and Dynamic Models (2Nd Ed.)*. Springer-Verlag New York, Inc., New York, NY, USA, 1997. ISBN 0-387-94725-6.

M. Zhou. Infinite edge partition models for overlapping community detection and link prediction. In *AISTATS*, pages 1135–1143, 2015.

M. Zhou. Discussion on "Sparse graphs using exchangeable random measures" by F. Caron and E. B. Fox. *Journal of the Royal Statistical Society: Series B (Statistical Methodology)*, 79(5):1359–1360, 2017.

M. Zhou and L. Carin. Negative binomial process count and mixture modeling. *IEEE Trans. Pattern Anal. Mach. Intell.*, 37(2):307–320, 2015.

M. Zhou, H. Chen, J. Paisley, L. Ren, G. Sapiro, and L. Carin. Non-parametric Bayesian dictionary learning for sparse image representations. In *NIPS*, 2009.

# Nonparametric Bayesian sparse graph linear dynamical systems: supplementary material

Rahi Kalantari, Joydeep Ghosh, and Mingyuan Zhou

## A Proof of Lemma 1

*Proof.* Since $z_{ij} \leq m_{ij}$ by construction, we have

$$\mathbb{E}[\|\mathbf{Z}\|_0] \leq \mathbb{E}\left[\sum_{i=1}^{\infty}\sum_{j=1}^{\infty} m_{ij}\right] = \mathbb{E}\left[\sum_{i=1}^{\infty}\sum_{j\neq i} r_i r_j + r_0 \sum_{j=1}^{\infty} r_j\right]$$

$$= \mathbb{E}\left[G^2 - \sum_{k=1}^{\infty} r_k^2 + r_0 G\right]. \quad (7)$$

Following Lemma 1 of Zhou (2015), it is straightforward to show that the right hand side of (7) is the same as that of (4) in main article. □

## B MCMC Convergence and Complexity Analysis

We show in Figure C.1 the trace plots of four representative model parameters, including two weight components of the gamma process $r_i$, the number of inferred nodes, and the total number of edges. The plots are obtained by running the model on the Beijing meteorological data. They show that the proposed Gibbs sampling algorithm converges fast and mixes well. Each Gibbs sampling iteration of the SGLDS has a complexity of $\mathcal{O}(KP^3 + K + N_Z + K^2 + TK^3)$, where $T$ is the length of observed time series, $K$ is the latent dimension of $\mathbf{x_t}$, $N_Z$ is the number of non-zero elements in the transition matrix ($\mathbf{W} \odot \mathbf{Z}$), and $P$ is the dimension of the observation. By contrast, a vanilla LDS has a sampling complexity of $\mathcal{O}(K + K^2 + TK^3 + KP^3)$. Considering that $N_Z < K^2$, we can conclude that our algorithm does not notably increase the complexity of the sampling algorithm.

## C Additional figures

Shown in Figure C.2 is the graphical representation of our model.

Figure C.3 shows that the loops within the inferred sparse random graph capture the seasonal components of the time series. Note different loops could have overlapping nodes.

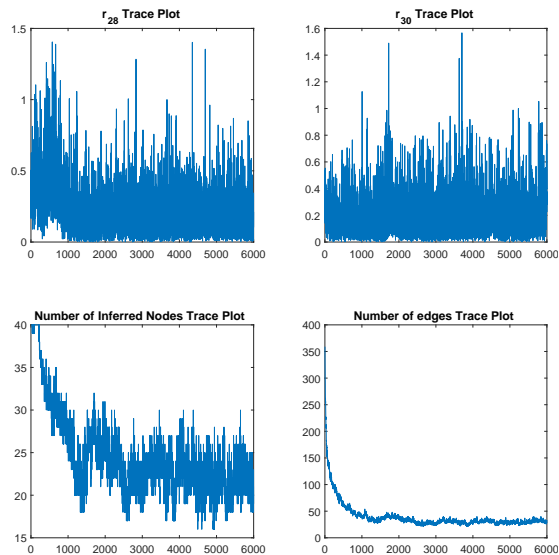

Figure C.1: Trace plots of four different model parameters.

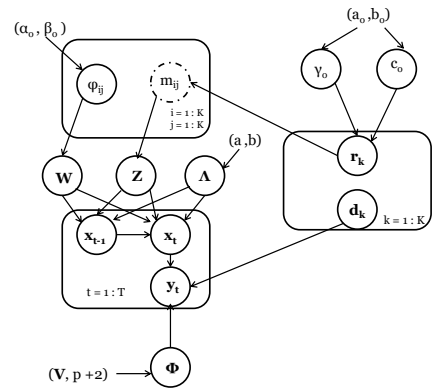

Figure C.2: Graphical Model

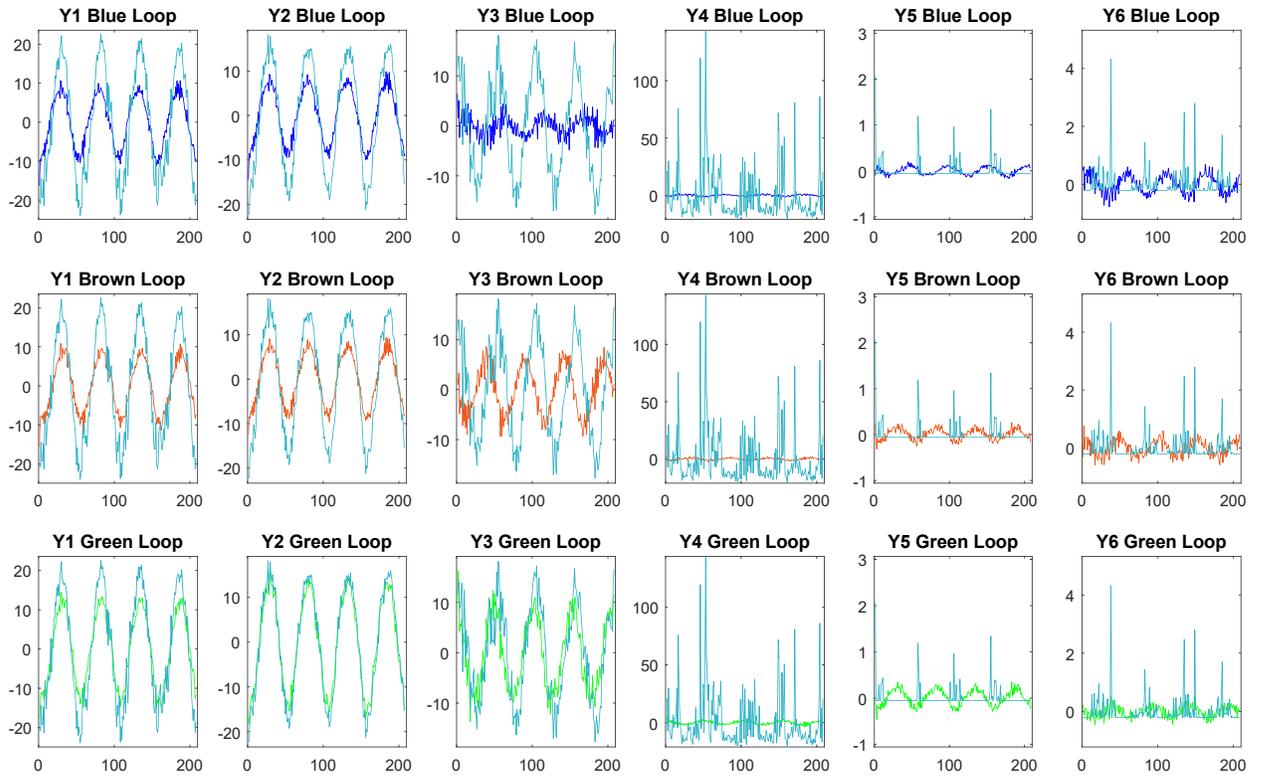

Figure C.3: The six-dimensional time series of the Beijing meteorological data and the reconstructed components using the states belonging to loop 1, 2, or 3 shown in Figure 2.